\documentclass[letterpaper, 10 pt, conference]{ieeeconf}  

\IEEEoverridecommandlockouts                              

\usepackage{fancyhdr}
\fancypagestyle{withfooter}{
  
  \fancyhead[L]{}
  \fancyhead[R]{}
  \fancyfoot[C]{\footnotesize Presented at the 2025 IEEE ICRA Workshop on Field Robotics}
}

\usepackage[l2tabu,orthodox]{nag}

\usepackage[
    backend=bibtex8,
    style=ieee,
    sorting=none,
    natbib=true,
    doi=false,
    isbn=false,
    url=false,
    eprint=false,
    maxcitenames=1,
    mincitenames=1
]{biblatex}

\usepackage[pdftex,colorlinks]{hyperref}

\usepackage[printonlyused]{acronym}

\usepackage{siunitx}
\usepackage[all]{nowidow}

\usepackage[dvipsnames]{xcolor}

\usepackage{lipsum}

\usepackage{xspace} 
\newcommand{\ie}{i.e.,\xspace{}}
\newcommand{\eg}{e.g.,\xspace{}}

\usepackage[pdftex]{graphicx}

\usepackage{epstopdf}

\usepackage{import}

\graphicspath{{./latexGoodPractices/}}

\usepackage{booktabs}

\usepackage{tabularx}
\usepackage{multirow, multicol}

\usepackage{amssymb,amsfonts,amsmath,amscd}

\usepackage{bm}

\usepackage[french,english,noabbrev,nameinlink]{cleveref}

\newcommand{\bbm}{\begin{bmatrix}}
\newcommand{\ebm}{\end{bmatrix}}

\usepackage{xcolor}
\usepackage{subcaption}
\usepackage{stfloats}

\usepackage{tabularray}
\usepackage{makecell}
\usepackage{siunitx}

\usepackage{lipsum}

\definecolor{asphalt}{HTML}{808080}
\definecolor{ice}{HTML}{0000FF}
\definecolor{gravel}{HTML}{FFA500}
\definecolor{grass}{HTML}{008000}
\definecolor{sand}{HTML}{FF4500}
\definecolor{mud}{HTML}{B8860B}
\definecolor{tile}{HTML}{F08080}

\acrodef{UGV}[UGV]{Unmanned Ground Vehicle}
\acrodef{IMU}[IMU]{Inertial Measurement Unit}
\acrodef{GNSS}[GNSS]{Global Navigation Satellite System}
\acrodef{DOF}[DOF]{Degrees of Freedom}
\acrodef{ICP}[ICP]{Iterative Closest Point}
\acrodef{SAR}[SAR]{Search-And-Rescue}
\acrodef{TnR}[TnR]{Teach-and-Repeat}
\acrodef{SLAM}[SLAM]{Simultaneous Localization and Mapping}
\acrodef{GUI}[GUI]{Graphical User Interface}
\acrodef{DDS}{Data Distribution Service}

\title{\LARGE \bf ASAP-MO: 
Advanced Situational Awareness and Perception for Mission-critical Operations
}

\author{Veronica Vannini$^{1}$,  William Dubois$^{1}$, Olivier Gamache$^{1}$, Jean-Michel Fortin$^{1}$, \\ 
Nicolas Samson$^{1}$, Effie Daum$^{1}$, François Pomerleau$^{1}$, Edith Brotherton$^{1}$
\thanks{$^{1}$Northern Robotics Laboratory, Université Laval, Québec
City, Canada, \textbraceleft {\tt\small veronica.vannini, william.dubois, olivier.gamache,  jean-michel. fortin, nicolas.samson, effie.daum, francois.pomerleau, edith.brotherton } \textbraceright
        {\tt\small
        @norlab.ulaval.ca}\
        }%
}

\begin{document}

\maketitle
\thispagestyle{withfooter}
\pagestyle{withfooter}

\begin{abstract}
Deploying robotic missions can be challenging due to the complexity of controlling robots with multiple degrees of freedom, fusing diverse sensory inputs, and managing communication delays and interferences. 
In nuclear inspection, robots can be crucial in assessing environments where human presence is limited, requiring precise teleoperation and coordination. 
Teleoperation requires extensive training, as operators must process multiple outputs while ensuring safe interaction with critical assets. 
These challenges are amplified when operating a fleet of heterogeneous robots across multiple environments, as each robot may have distinct control interfaces, sensory systems, and operational constraints. 
Efficient coordination in such settings remains an open problem.
This paper presents a field report on how we integrated robot fleet capabilities~\textemdash including mapping, localization, and telecommunication~\textemdash toward a joint mission.
We simulated a nuclear inspection scenario for exposed areas, using lights to represent a radiation source. 
We deployed two \acp{UGV} tasked with mapping indoor and outdoor environments while remotely controlled from a single base station. 
Despite having distinct operational goals, the robots produced a unified map output, demonstrating the feasibility of coordinated multi-robot missions. 
Our results highlight key operational challenges and provide insights into improving adaptability and situational awareness in remote robotic deployments.

\end{abstract}

\section{INTRODUCTION}

Years of training are required to master the teleoperation of a single robot, and performance significantly impacts the effectiveness of teleoperation \cite{baranski2006hfm136}.
On one side, an operator needs to fuse multiple artificial sensory inputs, possibly corrupted by communication delays and interferences, to understand the scene around the robot.  
On the other side, robots have multiple degrees of freedom, and the forces they produce can easily damage critical assets if not properly controlled. 

The challenge of teleoperation is amplified beyond human capabilities when a fleet of different robots is required for a task. 
For example, a nuclear decommissioning operation may require a variety of robotic capabilities on an indoor mission. 
Such a mission could include a maneuverable robot that can navigate through doorways and climb stairs, an articulated arm for precise manipulation of the environment, an aerial drone for visual inspections, and specialized robots equipped with specific sensors for hazard mapping~\cite{SimonWatsonNuclear}. 
Another context could be a ground \ac{SAR} operation in rough terrain that would require a heavy transport robot to quickly reach an incident zone (outside) and deploy a smaller maneuverable robot to access a building (inside) while passing doorways and climbing stairs, along with using an arm for fine manipulation of the environment. 
The recent DARPA Subterranean Challenge gathered worldwide laboratories in an underground environment and demonstrated how hard it can be to oversee a fleet of robots, with the winning team deploying walking robots \cite{Ebadi2023}. 

\begin{figure}[t]
    \includegraphics[width=\linewidth]{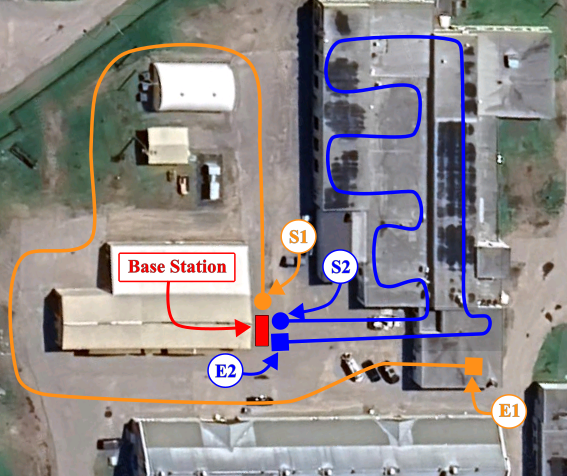}
    \centering
    \caption{Illustration of a mission. Where \textbf{(S1)} and \textbf{(E1)} represent the start and end route of the first robot, that scouts the outside environment, and \textbf{(S2)} and \textbf{(E2)}, show the plan start and end of the second robot, that accomplish the detection task. The full operation is controlled from the base station highlighted in red. This aerial view captures the real-world setting where the final demonstration was executed.}
    \label{fig:general}
\end{figure}


The 2011 earthquake and tsunami exposed the Fukushima Daiichi Nuclear Power Plant to severe radioactive conditions. 
The surveillance missions required modifying existing robots to enhance their radiation tolerance, mobility, and communication systems for inspecting the damage.
Some key challenges in this hazardous environment included ensuring the hardware's reliability under high radiation levels and training operators to use the robots effectively in simulated environments. 
Successful operations highlighted the importance of rigorous testing, custom adaptations for environmental conditions, and effective operator training. 
These factors are crucial for maximizing the robots' capabilities and ensuring safety during disaster response missions~\cite{nagatani2013}. 

We aimed to investigate whether recent advances in robotics were robust enough to enable a single operator or a small control team to conduct an inspection mission across indoor and outdoor environments. 
Therefore, our general objective was to explore the possibilities enabled by mobile robotics to help operators fulfill time-critical missions using one or more robots, depending on the tasks to be performed.
As illustrated in \autoref{fig:general}, our final deployment consisted of simulating a radiation decommission mission.   
A remote base station, set up in a secure environment, enables the operator to control the first robot to navigate a specified outdoor area, illustrated by the orange path, while mapping the surroundings. Afterward, a smaller robot designed for improved mobility is deployed to carry out a radiation detection task\textemdash~simulated through lights\textemdash traveling the blue path in the figure.

This field report presents how our multidisciplinary team was able to field test various situations (\eg~day-night, inside-outside, urban-forest) and successfully prepare a technical demonstration with four platforms (\ie~mobile control station, large \acf{UGV}, medium \ac{UGV}, and an articulated robotic arm) jointly used in an inspection mission while specifically addressing the following specific objectives:

\begin{enumerate}
    \item Developing a hierarchical \ac{SLAM} system that is common to a heterogeneous fleet and adaptable to the precision requirement of each robot to increase the situational awareness of an operator.  
    \item Exploring opportunities for new techniques or methods to reduce the manual interventions by operators, aiming toward increasing the autonomous tasks assigned to mobile robots for a given context and environment. 
    \item Demonstrating the capability of sensor data analysis techniques to detect specific environmental values, enhancing situational awareness, and introducing a new perception layer for the operator. 
\end{enumerate}

\section{METHODOLOGY}
\begin{figure*}[t]
  \includegraphics[width=0.99\textwidth]{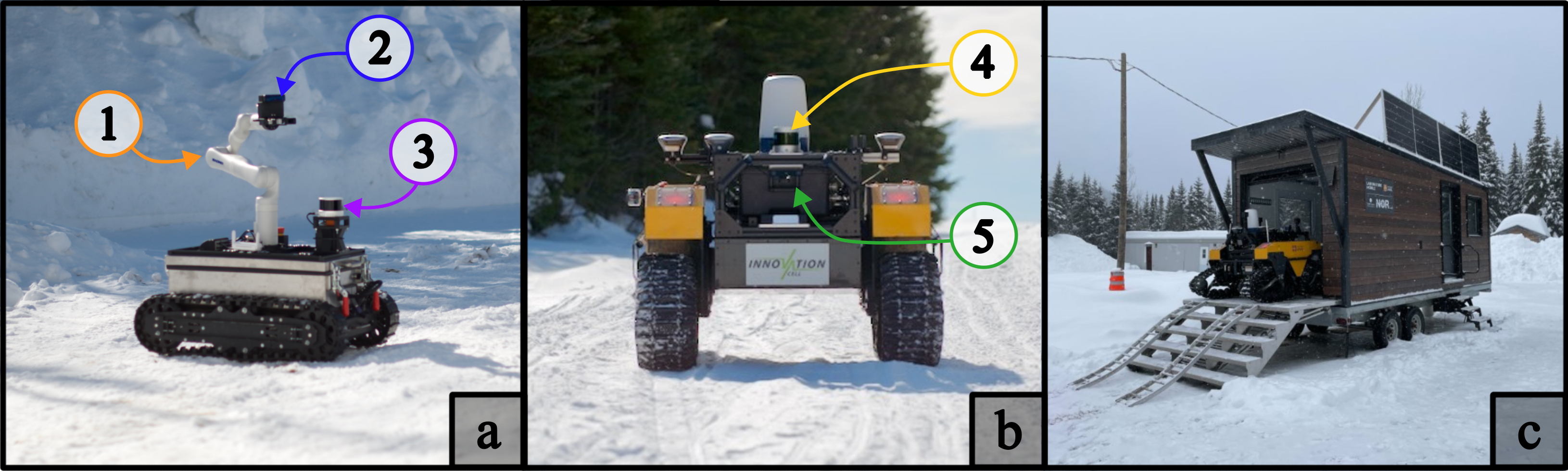}
  \centering
  \caption{Figure \textbf{a} shows the medium platform, HD2, equipped with: \textbf{(1)} Kinova Gen3 robotic arm, \textbf{(2)} the OAK camera and \textbf{(3)} Hesai XT-32 lidar. Figure \textbf{b} shows the Warthog from Clearpath, \textbf{(4)} Lidar RS128  and  \textbf{(5)} ZED X stereo camera. Figure \textbf{c} shows the mobile base station, to transport and operate the devices during experiments.}
  \label{fig:plarforms}
\end{figure*}

Experiments for this project were conducted in three different locations in a setting that blended outdoor and indoor areas. 
The operational range extended approximately \SI{100}{\meter} from the base station to the furthest explored point. 
The outdoor conditions were characterized by winter elements, including snow and snowbanks, which presented challenges for the robots' traction and navigation \cite{Pomerleau2023}. 
The indoor environment posed navigation challenges due to walls, doors, and diverse obstacles, which the robots had to navigate around or through.

Three adaptable robotic platforms were chosen for the experiment, each with unique capabilities. 
The medium-sized platform, \autoref{fig:plarforms}\textcolor{red}{a}, is an HD2 treaded tank platform from Superdroid Robots, weighing approximately \SI{65}{\kilo\gram}, and equipped with a Hesai XT-32 lidar and an XSens MTi-10 \ac{IMU}. 
A Kinova Gen3 robotic arm with \num{7} degrees of freedom was integrated on top of the HD2 to expand its functionality. 
Robotic arms are widely used to address mobility limitations \cite{Fall2015}, and in this deployment, the Kinova arm provided increased versatility. 
A Luxonis OAK-D Lite camera was mounted at the end of the arm, giving the operator greater control over the camera and different scanning angles for better visibility of the environment. 
Finally, the Warthog, \autoref{fig:plarforms}\textcolor{red}{b}, a ground robot from Clearpath, weighing approximately \SI{500}{\kilo\gram}, was modified to be equipped with a Navtech CIR304-H radar, a ZED X stereo camera for teleoperation, as well as two high resolution lidars: the RoboSense RS-Ruby Plus, a 128-beam LiDAR (RS128) designed for long-range, high-resolution mapping, and the Leishen LS-S1 (LS128), a hybrid solid-state LiDAR optimized for precise obstacle detection and navigation. 
The Warthog also featured an XSens MTi-10 \ac{IMU}, a VN100 \ac{IMU}, and three Emild Reach GNSS units. 
Our stationary base station, to facilitate centralized teleoperation and monitoring throughout the experiments, is shown in \autoref{fig:plarforms}\textcolor{red}{c}.
Our initial experiments were executed on the line of sight from the range of the base station (not teleoperated) using a standard \SI{5}{\giga\hertz} network.  
This configuration was upgraded to three Multiband Wearable Mesh Rider Radios from Doodle Labs shortly before the final demonstration. 
The upgraded configuration with the Doodle Labs Radios allowed us to teleoperate with all robotic assets from a centralized base station while creating a unified global map containing outdoor and indoor environments. 
By using one radio per platform, forming a mesh network between the devices, we were able to teleoperate within approximately \SI{100}{\meter}.

\begin{figure}[htbp]
    \includegraphics[width=1.0\linewidth]{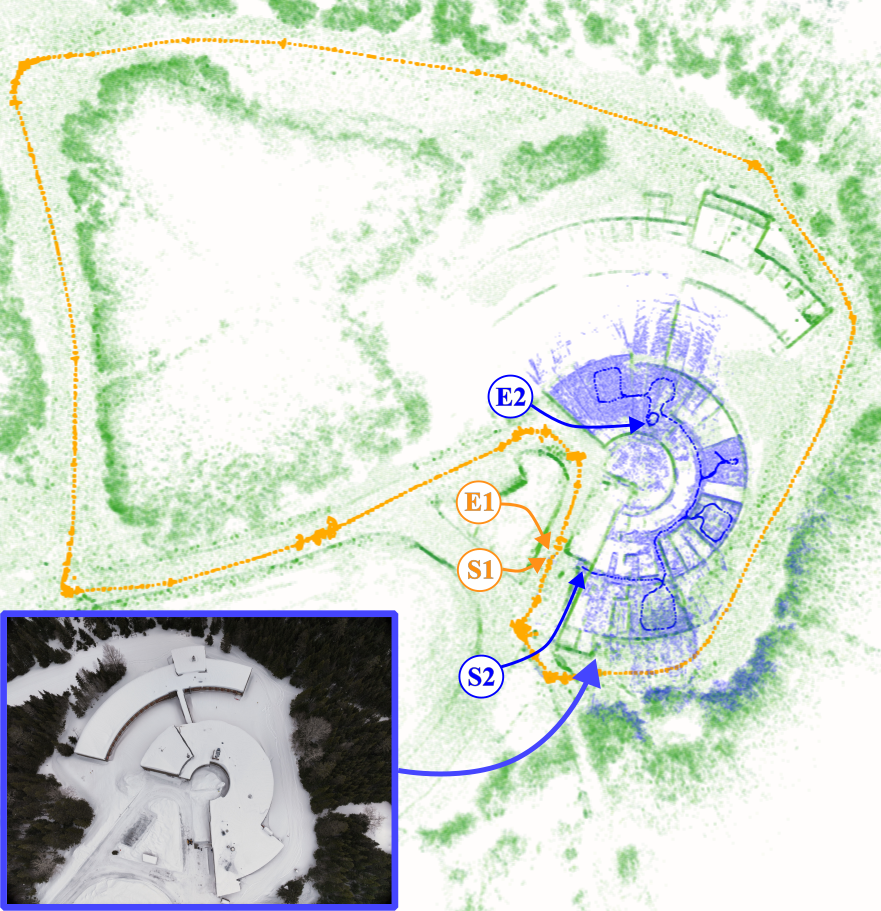}
    \centering
    \caption{The green map and orange trajectory represent the results from one of the Warthog’s runs at the Forêt Montmorency (starting at S1 and ending at E1). The blue map consists of additional points collected by the HD2 as it followed the blue trajectory (from S2 to E2), which is shown in greater detail in \autoref{fig:castor_map}.}
    \label{fig:full_map}
\end{figure}

\begin{figure}[htbp]
    \includegraphics[width=\linewidth]{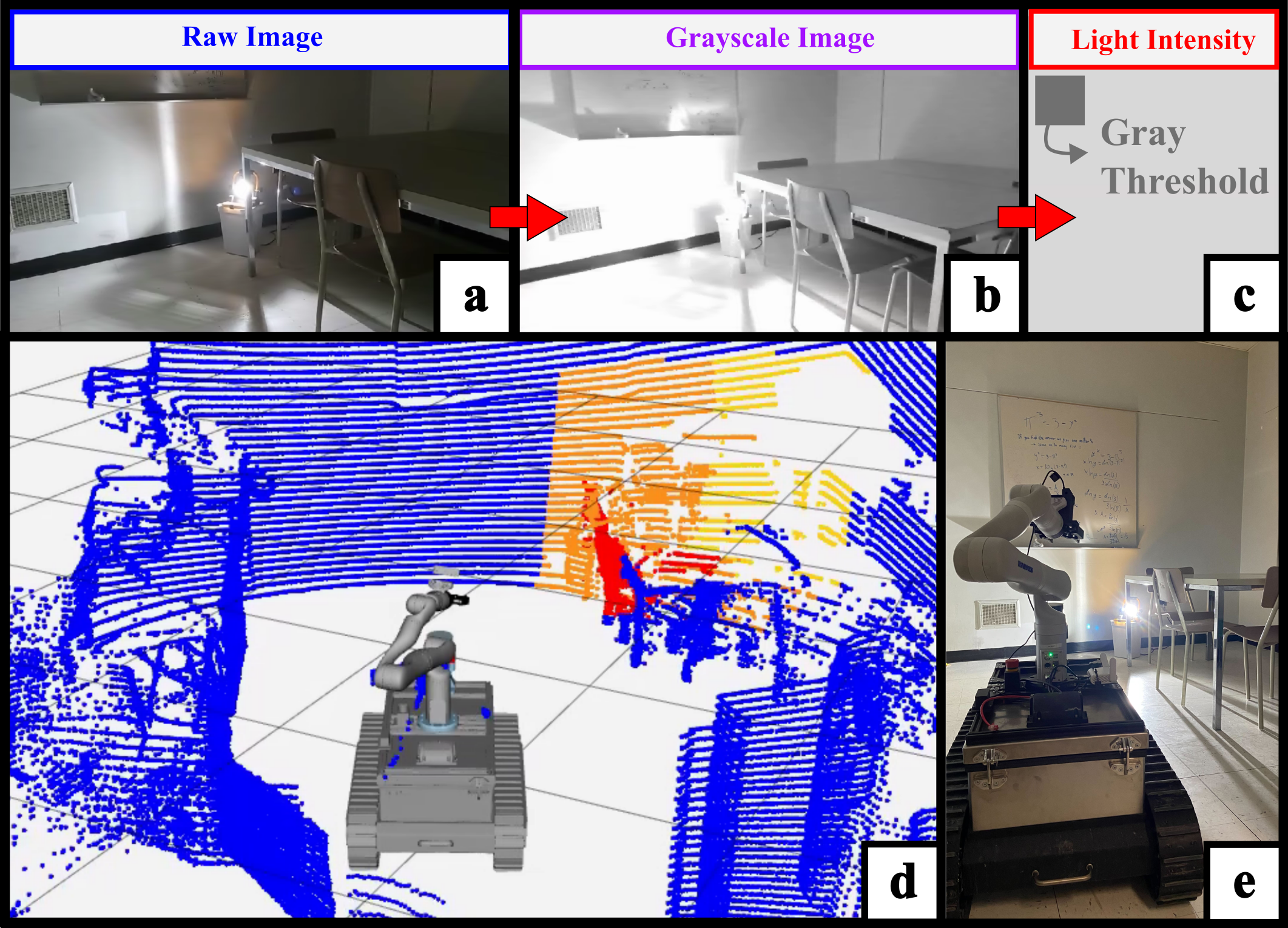}
    \centering
    \caption{The process to convert a camera image to a directional Geiger counter. Identify and add light intensity to the map. 
    Figure \textbf{a} shows the image using auto-exposure used by the operator to see the environment. 
    Figure \textbf{b} shows the same view in grayscale with a fixed exposure used to mock the Geiger counter (i.e., not accessible to the operator).
    Figure \textbf{c} presents the average grayscale value retrieved from the image and the comparison with the grayscale threshold. 
    Figure \textbf{d} shows the projection to the map. 
    The red, orange, and yellow points meet the light intensity criteria and have two, three, and four meters of distance, respectively. 
    Figure \textbf{e} shows a picture of the real environment at Université Laval.}
    \label{fig:light_detction}
\end{figure}

To achieve our objectives, all field experiments are conducted following a structured, multi-stage methodology outlined as follows: \textbf{(Step 1)}
The Warthog is launched from the base station to map the building's exterior and find a safe entrance inside.
While teleoperating, the resulting initial outdoor map, as shown in green in \autoref{fig:full_map}, is displayed to the operator. 
\textbf{(Step 2)} The initial outdoor map is then transferred to the HD2 via the network. 
As the transfer is wireless, the robots only need to be within operational range to connect, allowing robots to be in different starting and ending locations, as \textbf{S2} and \textbf{E1} in \autoref{fig:general}. 
\textbf{(Step 3)} The map received by the HD2 platform has different initial points than its own, making it necessary to relocalize the robot to the right spot on the map. 
The operator needs to input the approximated position on the map in the \ac{GUI} for the system to find the right location and properly continue to map, allowing indoor exploration while maintaining a consistent global reference. 
\textbf{(Step 4)} With a \emph{follow the gap} algorithm \cite{SEZER20121123}, enhanced with a heading incentive to navigate the unfamiliar environment effectively, the HD2 is teleoperated from the base station to inside the building until the operator reaches an area of interest.
\textbf{(Step 5)} At any point during the trajectory of the HD2 robot, the operator can move the robot arm to perform a detailed area inspection and search for radioactive areas.
For safety reasons, lights are used as a proxy for radiation sources (\ie~bright construction lights) as shown in \autoref{fig:light_detction}.
The OAK camera captures a raw image (\autoref{fig:light_detction}\textcolor{red}{a}), using a fixed exposure configuration and converts it to a grayscale image (\autoref{fig:light_detction}\textcolor{red}{b}). 
A single value, computed as the average of all grayscale pixels, represents the light intensity and is displayed, as shown in \autoref{fig:light_detction}\textcolor{red}{c}. 
\textbf{(Step 6)} When the light intensity is greater than \SI{44}{\percent} (\ie~112 in 8-bit image), the area is considered radioactive and this sensor data is projected onto the map.
This threshold is particular to the combination of lights and camera at hand, but was adjusted to simulate a directional Geiger counter.
We assigned a value based on the distance from the robot to each matching point in the map projected by the camera.  
\autoref{fig:light_detction}\textcolor{red}{d} shows the differences in values for the same reading, where red points are \SI{2}{\meter} or closer, orange points are between \SI{2}{\meter} and \SI{3}{\meter}, and yellow points are at least \SI{4}{\meter}. 
The color represents how strong the radiation would be detected by the Geiger counter.
Blue points are the ones not seen by the camera. 
As new observations are made, previous values are updated to refine the precision of the radiation distribution map. 

\begin{figure}[htbp]
    \includegraphics[width=\linewidth]{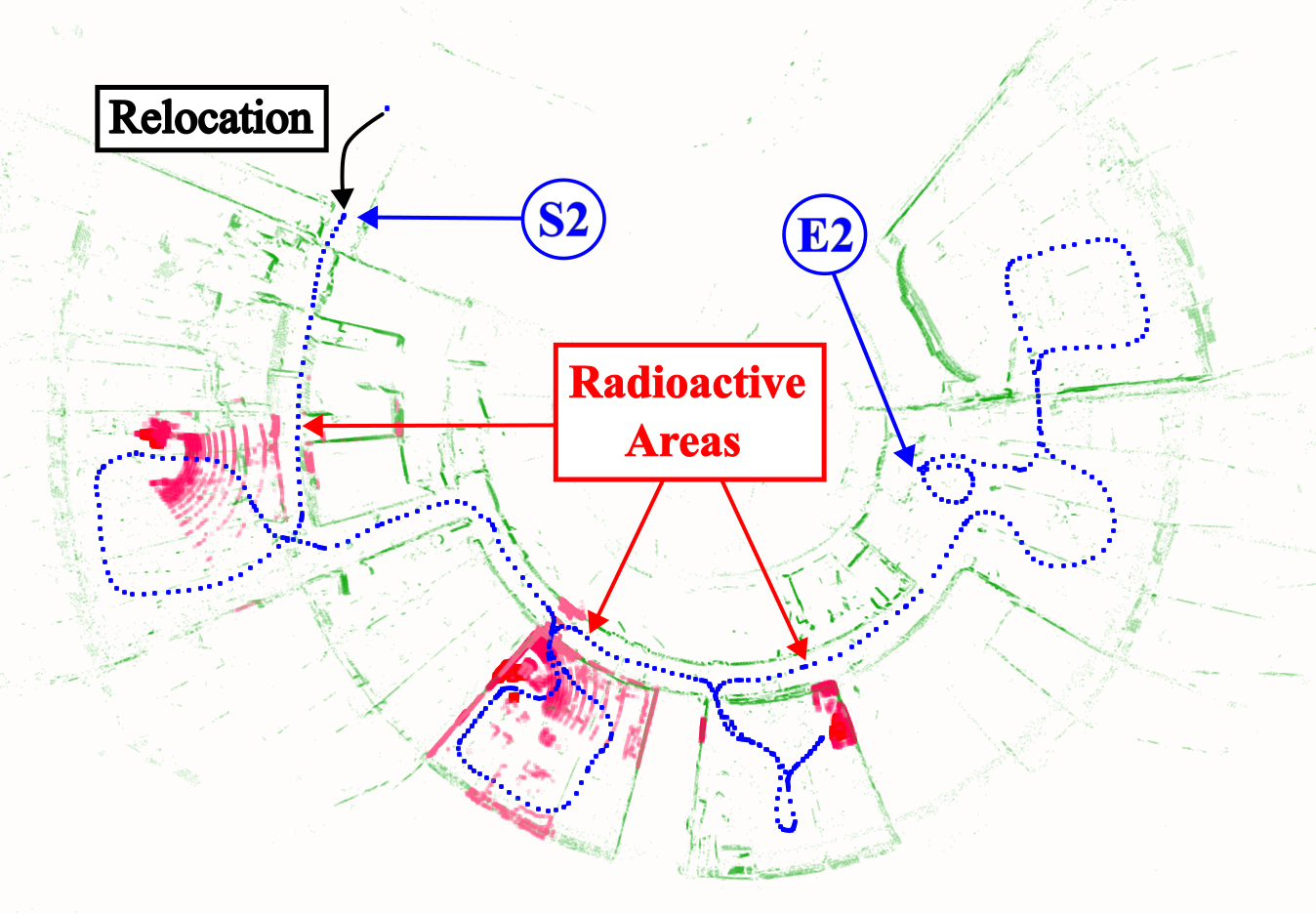}
    \centering
    \caption{Indoor mapping from the HD2 robot, with the initial relocation in the Warthog map, the trajectory in blue, and discovered radioactive areas in red.}
    \label{fig:castor_map}
\end{figure}

The \ac{ICP} algorithm~\cite{Pomerleau12comp} was used for all mapping in the project. 
For the HD2 robot, we integrated additional descriptors for light intensity to enhance our world representation on the map and increase perception and awareness of the operator.
\autoref{fig:castor_map} shows the robot's trajectory while mapping inside the main building at Montmorency Forest.
The points highlighted in red show where the robot found radiation sources (i.e., exposed areas).

 \autoref{table:lessons} presents a summary of all experiments conducted, with the type of robots (\ie~Warthog, HD2), whether it was teleoperated or not, the time of day, the light detection algorithm (\ie~none, post processed, real-time), and finally the network configuration (\ie~\SI{5}{\giga\hertz}, \SI{915}{\mega\hertz}). 
 The location where each experiment took place is presented as UL for University Laval, FM for Montmorency Forest, and Final for our final demonstration at Valcartier Research Centre, where we had the opportunity to demonstrate our findings and challenges to soldiers and defense stakeholders.
 The table also provides quantitative metrics for total distance traveled by each robot, the duration of the experiments, and the average physical area covered. 
 Our experiments lasted over \SI{5.6}{\hour} for a total of \SI{5.86}{\kilo \meter}. 
 As we can observe, the HD2 platform was the most used with a total distance of \SI{3.24}{\kilo \meter}, in comparison with the Warthog at \SI{2.62}{\kilo \meter}. 
 Most of experiments were done with an operator driving the platform with visual line of sight (\SI{68}{\percent}) and in daylight (\SI{69}{\percent}) to integrate progressively the algorithms and assess the feasibility of deploying such robotic systems in larger or more complex environments.

\begin{table*}[htbp] 
\centering
\caption{Overview for all experiments.}
\begin{tblr}{Xcccccccc} 
\cline[1pt, black]{-}
\textbf{Platform} & \textbf{Local} & 
\makecell{\textbf{Teleoperated}} & \makecell{\textbf{Time of the Day}} & \makecell{\textbf{Light Detection}}  & \textbf{Network} & \makecell{\textbf{Sum Distance} \\\textbf{travelled (m)}} & \makecell{\textbf{Sum} \\\textbf{time (min)}} &
\makecell{\textbf{Average Physical space} \\\textbf{(\SI[detect-weight=true,mode=text]{}{\square \meter})}} \\ 
\cline[1pt, black]{-}
HD2 & \textcolor{sand}{UL} & - & day & - & \SI{5}{\giga\hertz} & 60 & 15 & 200 \\
HD2 & \textcolor{sand}{UL} & - & day & Real Time & \SI{5}{\giga\hertz} & 425 & 52 & 314 \\
HD2 & \textcolor{sand}{UL} & - & day & Post Processed & \SI{5}{\giga\hertz} & 240 & 24 & 200 \\
HD2 & \textcolor{asphalt}{FM} & - & night & Post Processed & \SI{5}{\giga\hertz} & 1530 & 50 & 4000 \\
HD2 & \textcolor{asphalt}{FM} & - & night & Real Time & \SI{5}{\giga\hertz} & 60 & 9 & 1800 \\
HD2 & \textcolor{asphalt}{FM} & - & night & - & \SI{915}{\mega\hertz} & 260 & 9.5 & 4000 \\
HD2 & \textcolor{asphalt}{FM} & Y & night & Post Processed & \SI{915}{\mega\hertz} & 330 & 27.5 & 1750 \\
HD2 & \textcolor{grass}{Final} & - & day & Real Time & \SI{915}{\mega\hertz} & 30 & 2 & 275 \\
HD2 & \textcolor{grass}{Final} & Y & day & Post Processed & \SI{915}{\mega\hertz} & 305 & 21.75 & 2875 \\
\cline[1pt, black]{-}
Warthog & \textcolor{sand}{UL} & - & night & - & \SI{5}{\giga\hertz} & 380 & 6 & 57000 \\
Warthog & \textcolor{asphalt}{FM} & - & day & - & \SI{5}{\giga\hertz} & 690 & 34.25 & 14333 \\
Warthog & \textcolor{asphalt}{FM} & - & day & - & \SI{915}{\mega\hertz} & 685 & 26 & 13250 \\
Warthog & \textcolor{grass}{Final} & Y & day & - & \SI{915}{\mega\hertz} & 860 & 56.25 & 16900 \\
\cline[1pt, black]{-}
\end{tblr}
\label{table:lessons}
\end{table*}

\section{CHALLENGES AND LESSONS LEARNED}

In this section, we outline the main challenges faced during deployments and share the lessons learned in these environments, aiming to assist future multi-staged missions involving a multi-robot fleet in real-world environments. 

\subsection{Networking}
\begin{figure}[htbp]
    \includegraphics[width=\linewidth]{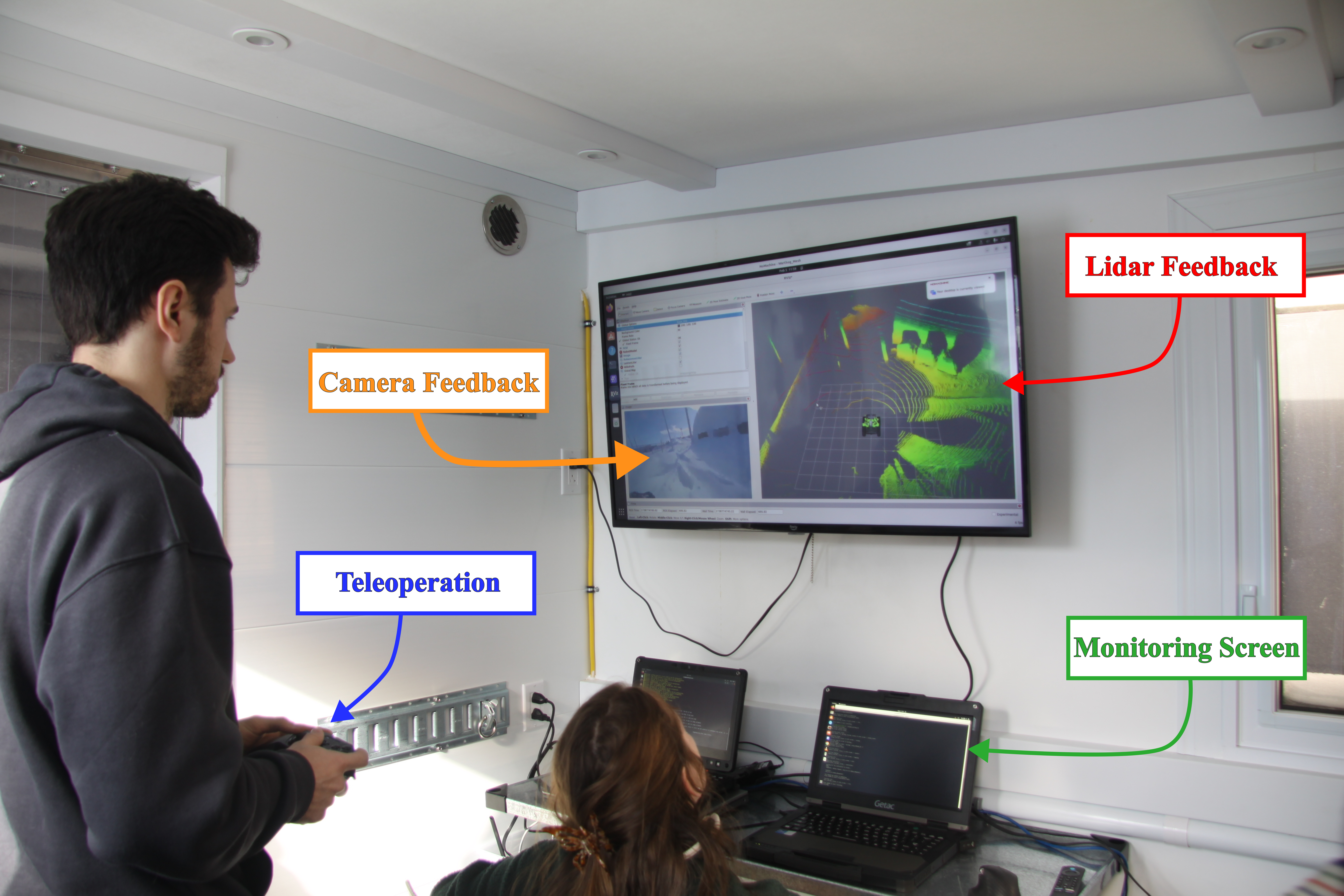}
    \centering
    \caption{Teleoperation tools installed in the base station.
    The rugged laptop is used to send commands to the robots, while the monitor allows the operator a larger visualization of the robots' main modalities for teleoperation, \ie~3D point cloud and camera image stream.}
    \label{fig:control}
\end{figure}

Networking and communication were important components of a successful experiment. 
To effectively teleoperate, it was necessary to visualize both the point cloud from the map and the camera feed at the base station, as shown in \autoref{fig:control}.
To achieve a reliable communication with the base station, there are mainly two aspects to take into account: the hardware used and the communication protocol.
For the hardware solution, a \SI{5}{\giga\hertz} router network was initially set up to create a mesh configuration using Ubiquiti UAP-AC-M radios to improve communication between the robots and the base station. 
However, the necessity of setting up the mesh network before the experiment is a limitation for a mission in hazardous environments, such as a nuclear decommissioning operation. 
Furthermore, the high frequency of the \SI{5}{\giga\hertz} signal resulted in significant challenges when operating indoors, as it struggled to penetrate walls effectively, leading to communication issues with the HD2 while inside buildings. 
Additionally, this network was susceptible to interference from other Wi-Fi signals within the building, further aggravating connectivity problems.

To overcome these challenges, we updated our network from the Ubiquiti hardware to the Doodle Labs Radios, with a lower-frequency \SI{915}{\mega\hertz} signal, allowing for more effective wall penetration and a more extended range. 
A mesh network was established between the base station, Warthog, and HD2 antennas, enabling any platforms to be strategically positioned as communication relays in the field. Yet, the lower bandwidth of this configuration imposed restrictions on raw data transmission.
Based on our limited throughput, two possible communication protocols were selected for investigation, being using the Zenoh \ac{DDS} with ROS2, or a remote desktop-based solution.
Each computer involved was configured with ROS2 Humble and with Zenoh as communication protocol, where each machine was a Zenoh Router.
This first setup showed great promises for wireless communication at close range, or with small message types, such as velocity commands.
Unfortunately, passing images and 3D point clouds only worked with line-of-sight communication, which does not follow our need for multi-environments navigation.
Consequently, the final solution only used Zenoh to send the teleoperation commands throughout ROS2 on each robot.
The second setup that was tested is based on a remote desktop solution named NoMachine NX.
By installing NoMachine NX on each platforms, the raw camera stream and 3D map did not have to pass throught the network since only the screen visualization was displayed to the  base station.
It was used for remote access to the onboard computers, ensuring low-latency communication while facilitating efficient data transfer and mitigating the data transmission constraint.

\subsection{Teleoperation}
One key lesson learned regards the system's overall usability, including its commands and the importance of understanding the mission's objectives alongside the robots' mechanics. 
Controlling the system from a single base station simplifies the process. 
However, it introduces additional operational complexity for the user, who must be familiar with all the robots and verify all configurations when switching control from one platform to another. 
Furthermore, users should be familiar with teleoperation, as there may be some delay in the data link between the camera, lidar, and the base station. 
The user must remain aware and take appropriate actions to mitigate risks to the robot and its environment, mainly when operating in manual mode. 
The operator can also use some other semi-autonomous techniques and methods to mitigate risks, such as following the gap mode or implementing a \ac{TnR} algorithm~\cite{Baril2022}, so the robot can autonomously return to the starting point using the previously recorded path.

\begin{figure}[htbp]
    \includegraphics[width=\linewidth]{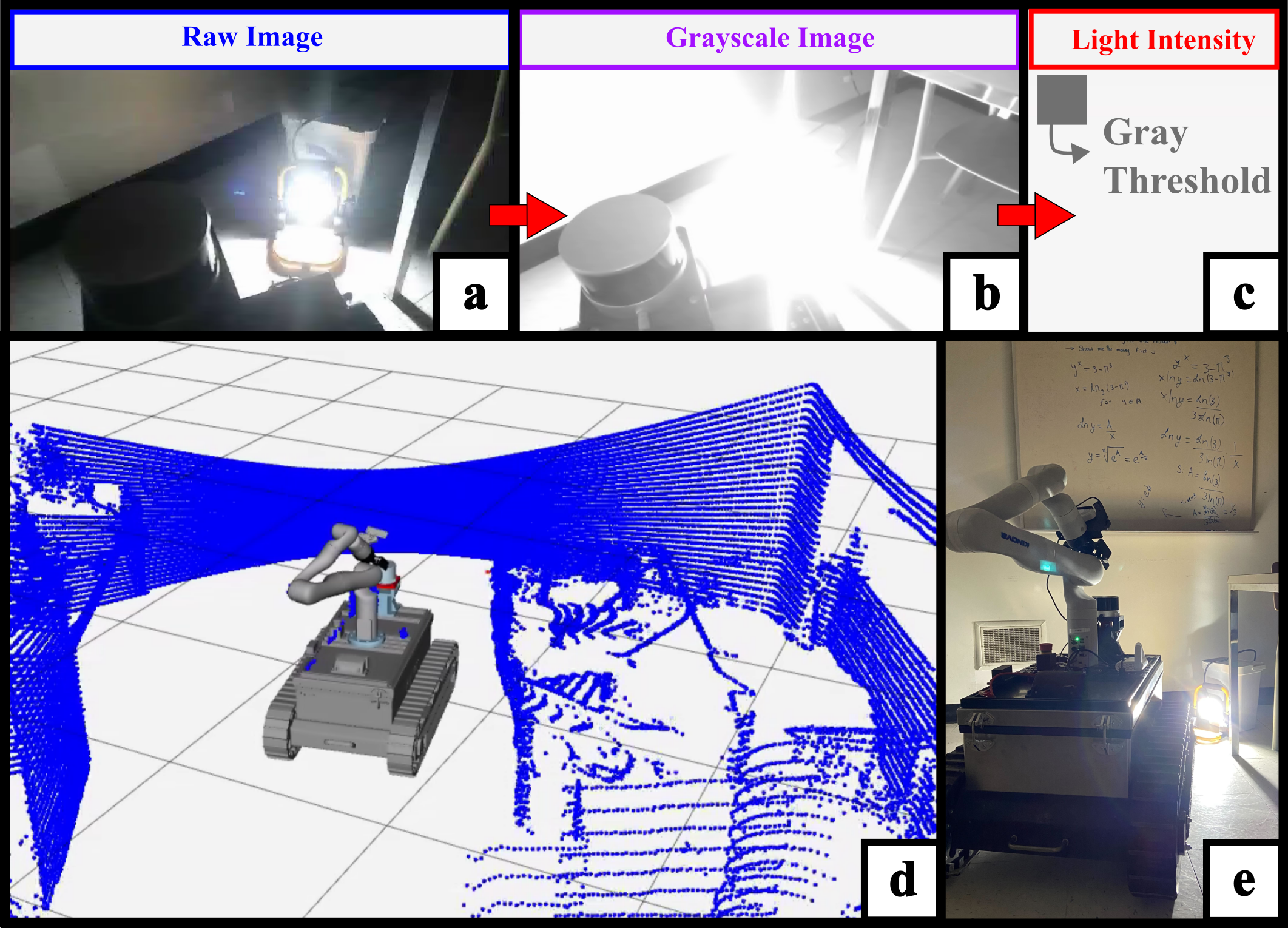}
    \centering
    \caption{Example of a misleading environmental reading. Figure \textbf{a} presents the raw camera image with auto exposure, Figure \textbf{b} shows its grayscale conversion with fixed exposure, and Figure \textbf{c} visualizes the detected light intensity, all confirming that the spotlight is within the camera's field of view. However, Figure \textbf{d} shows the lidar points with only blue points, as none of the camera readings can be projected into the current points. Figure \textbf{e} shows a picture of the actual environment at Université Laval.} 
    \label{fig:error_light}
\end{figure}

Regarding the mission's context and the robot's limitations, the operator should be aware of the different points of view between the camera that detects light sources and the position of the lidar. 
When the robot is close to a radiation source on the ground, and the lidar is oriented parallel to the ground, the system is unable to accurately project the camera readings, as the field of view doesn't overlap with the lidar scan, resulting in missing detections. 
Therefore, the operator needs to understand the system's limitations to achieve results that accurately reflect reality. 
This issue could be resolved by projecting light data onto the map points instead of using the lidar readings; however, this method was too time-consuming to be performed in real-time in our case. 
\autoref{fig:error_light} illustrates a recorded situation where the camera can observe the light source, but the information can not be translated to the map.

\section{DISCUSSIONS AND CONCLUSION}

Our final demonstration to soldiers and defense stakeholders was a great opportunity to engage in conversations with operators having extended field knowledge about teleoperating a robot fleet in mission-critical operations. 
The conversation with stakeholder representatives was enlightening and helped us better understand their current configuration's limitations, which involves two robots (i.e., a large and a small platform) similar to our proposal.
These representatives had experience in real combat scenarios.
Mechanically, their robots are strong enough to go through gypsum walls and can climb stairs of almost any slope without flipping. 
One of the challenges they face is that their robots need to be wired with optical fiber and a winder to maintain communication with the base station. 
This physical connection allows them for a trajectory up to \SI{150}{\meter}, which is a similar range as the Doodle Labs Radios used in our project. 
They explained that our teleoperation solution, which operates without a tether, could significantly benefit their operations, as they often conduct tasks without direct visual contact with their robots and need to backtrack their way on a trajectory to be able to continue exploring.

The stakeholders' system offers no information regarding slopes or angles, resulting in a flat perception of the environment. 
Utilizing feedback from lidar and cameras, such as the one in our solution, would greatly improve tasks such as identifying soft obstacles like grass. 
Using feedback from the \ac{IMU} and other sensors can greatly improve traversability in a robotic mission and aligns with our laboratory's upcoming research activities~\citet{fortin2025}.
Another challenge the operators face in their missions is the limited space within the vehicle they use to transport the robots and conduct the missions from. 
Their vehicle can seat at most three individuals, who are usually a technical chief, a driver, and a drone operator, with minimal comfort and space.
These observations highlight that future design of inspection robots for time-critical tasks must carefully handle a balance between size and mechanical capability.


In this work, we successfully demonstrated heterogeneous platforms (\eg~mobile control station, large reach \ac{UGV}, short reach \ac{UGV}, and a robotic arm) working in a common framework to investigate the safety of a radioactive area.
Through this process, we gained valuable insights into the current opportunities and constraints associated with robotic systems in Canada, particularly by placing the robots in challenging scenarios and attempting to complete missions with minimal human intervention. 
Future research will concentrate on advancing perception, navigation, and motion capabilities under extreme environmental conditions such as snowstorms, black ice, fog, and flooding. 
This will include the development of algorithms capable of converting planned trajectories to accommodate various mobility modalities, including aerial navigation, terrestrial locomotion, and skid-steering mechanisms.

\addtolength{\textheight}{-12cm}   


\section*{ACKNOWLEDGMENT}

This research was supported by the Department of National Defence (DND), Defence Research and Development Canada (DRDC), Valcartier Research Centre (VRC), and more specifically, Benoit Ricard.


\printbibliography

\end{document}